\documentclass{article}
\usepackage[english]{babel}
\usepackage[latin1]{inputenc}
\usepackage{graphicx}
\usepackage[margin=0.9in, footnotesep=0.2in]{geometry}
\usepackage{pifont}
\usepackage{epstopdf}
\usepackage{epsfig}
\usepackage{subfig}
\usepackage{calc}
\usepackage{amssymb}
\usepackage{amstext}
\usepackage{comment}
\usepackage{amsmath} 
\usepackage{amsfonts}
\usepackage[square,sort,comma,numbers]{natbib}
\usepackage{multicol}
\usepackage{pslatex}
\usepackage{verbatim}
\usepackage{hyperref}
\usepackage[para]{threeparttable}
\usepackage{graphics}
\usepackage{setspace}
\usepackage[labelsep=period]{caption}
\usepackage{csquotes}

\usepackage{paralist}
\usepackage[title]{appendix}

\usepackage{url}

\renewenvironment{abstract}
 {\small
  \list{}{%
    \setlength{\leftmargin}{15mm}
    \setlength{\rightmargin}{\leftmargin}%
  }%
  \item[\textbf{\hspace{8mm} Abstract ---}]\relax}
 {\endlist}

\begin{document}
\tolerance=1600

%
\title{Predicting overweight and obesity in later life from childhood data: A review of predictive modeling approaches}
%
\author{Ilkka Rautiainen\footnote{\textbf{Correspondence:} ilkka.t.rautiainen@jyu.fi, P.O.Box 35, FI-40014 University of Jyvaskyla, Finland} \and Sami \"{A}yr\"{a}m\"{o}}
\date{}
\maketitle              

\vspace{-1cm}
\begin{center}
\small
Faculty of Information Technology, University of Jyvaskyla, Finland
\end{center}

\begin{abstract}
\textit{Background}: Overweight and obesity are an increasing phenomenon worldwide. Predicting future overweight or obesity early in the childhood reliably could enable 
a successful intervention by experts. While a lot of research has been done using explanatory modeling methods, capability of machine learning, and predictive modeling, 
in particular, remain mainly unexplored. In predictive modeling models are validated with previously unseen examples, giving a more accurate estimate of their 
performance and generalization ability in real-life scenarios.

\textit{Objective}: To find and review existing overweight or obesity research from the perspective of employing childhood data and predictive modeling methods. 

\textit{Methods}: The initial phase included bibliographic searches using relevant search terms in PubMed, IEEE database and Google Scholar. 
The second phase consisted of iteratively searching references of potential studies and recent research that cite the potential studies.

\textit{Results}: Eight research articles and three review articles were identified as relevant for this review. 

\textit{Conclusions}: Prediction models with high performance either have a relatively short time period to predict or/and are based on late childhood data. 
Logistic regression is currently the most often used method in forming the prediction models. In addition to child's own weight and height information, maternal 
weight status or body mass index was often used as predictors in the models. 


\end{abstract}

\section{Introduction}

Obesity is a global phenomenon that has increased rapidly during the last few decades in most countries. 
This trend has also led to significant increase in obesity-related diseases and deaths 
\cite{the_gbd_2015_obesity_collaborators:_health_effects_of_overweight_and_obesity}. 
The review presented here aims to map and review the methods used in overweight and obesity prediction research with an 
emphasis on predictive modeling techniques. 
The basic question to consider is ''Can we predict a person's overweight or obesity status in later 
life from the data collected during childhood?'' Ideally, early identification will make it possible to take steps for a 
successful obesity intervention. 
Currently this identification is often done manually by using growth references such as 
\cite{onis:_development_of_a_who_growth_reference}, 
\cite{saari:_new_finnish_growth_references_for_children_and_adolescents} and 
\cite{cole:_extended_international_iotg_bmi_cutoffs}.
Also, the data collection phase can be a tedious and expensive process. It is therefore preferred 
that this identification can be achieved with easily available basic data, such as height and weight information. 
A widely used measure derived from these two attributes is body mass index (BMI).
Ideally, children in unhealthy BMI trajectories should be identified before school age 
\cite{lynch:_development_of_distinct_bmi_trajectories_among_children_before_age_5_years}. 
For adults, the BMI cut-off points in widest use are 25 kg/m\textsuperscript{2} for overweight and 
30 kg/m\textsuperscript{2} for obesity \cite{cole:_establishing_a_standard_definition_of_childhood_obesity_worldwide}. Age and sex specific cut-off values 
that can be used in children have also been developed \cite{cole:_body_mass_index_reference_curves_for_the_uk, cole:_establishing_a_standard_definition_of_childhood_obesity_worldwide}.

When validating the employed prediction model, the performance measures are always a trade-off between sensitivity (in this case the ratio of 
correctly classified overweight or obese cases in relation to total overweight or obese cases) and specificity 
(the ratio of correctly classified normal weight cases in relation to total normal weight cases).
In this context, the most important performance measure is sensitivity, which indicates the proportion of 
children we are most interested in finding, in order to target preventive actions in this group 
\cite{zhang:_comparing_data_mining_methods_with_logistic_regression_in_childhood_obesity_prediction}. 
However, when the specificity is low, Butler et al. (2018) \cite{butler:_childhood_obesity__how_long_should_we_wait_to_predict_weight} 
argue that it becomes questionable to use the model at all, since the model generates large amounts of false positives. 
They argue further that the risk threshold should be placed based on considering multiple criteria, 
including potential risks and harms as well as financial costs. 
Additionally, if a suitable tool for obesity prediction became eventually available for clinical use, careful steps should be 
taken by the practitioners to avoid potential ethical issues in interventions. Communicating the overweight/obesity risk to child's parents might have undesirable 
and unforeseen consequences. Also, availability of practical remedies should be ensured for those deemed to be at significant risk 
for obesity \cite{levine:_identifying_infants_at_risk_of_becoming_obese__can_we_and_should_we}.

Breiman (2001) \cite{breiman:_statistical_modeling} and Shmueli (2010) \cite{shmueli:_to_explain_or_to_predict} have discussed the differences between explanatory and predictive types of modeling. 
Explanatory modeling is defined as 
''the use of statistical models for testing causal explanations'' and predictive modeling as ''the process of applying a statistical model or data mining 
algorithm to data for the purpose of predicting new or future observations'' \cite{shmueli:_to_explain_or_to_predict}. 
Validation is used in predictive modeling to estimate how well the model is expected to work with previously unseen data 
\cite{bishop:_pattern_recognition_and_machine_learning}. 
According to Bzdok et al. (2018) \cite{bzdok:_points_of_significance}, prediction makes it possible to identify the best courses of action without requirement to understand 
the underlying mechanisms. 
In explanatory modeling the model is not validated with an independent test data and thus its 
performance in the task of predicting new observations might be overestimated \cite{sainani:_explanatory_versus_predictive_modeling}. 

\section{Material and methods}\label{sec:material_and_methods}

The scope of this review included studies in English language that:
\begin{enumerate}
 \item Predict the future overweight/obesity status with a model built using the baseline data collected in childhood (e.g. using the data collected up 
       to age three years to predict child's obesity status at age six years).
 \item In addition to describing the prediction framework also report relevant numeric results (e.g. sensitivity and specificity and/or AUC values).
 \item Validate the results using an internal and/or external independent test set that has not been used for training the model. 
       Internal bootstrap validation is not considered here to be an independent test data set.
\end{enumerate}

\begin{figure}[h]\centering
  \includegraphics[height=5.3cm,keepaspectratio]{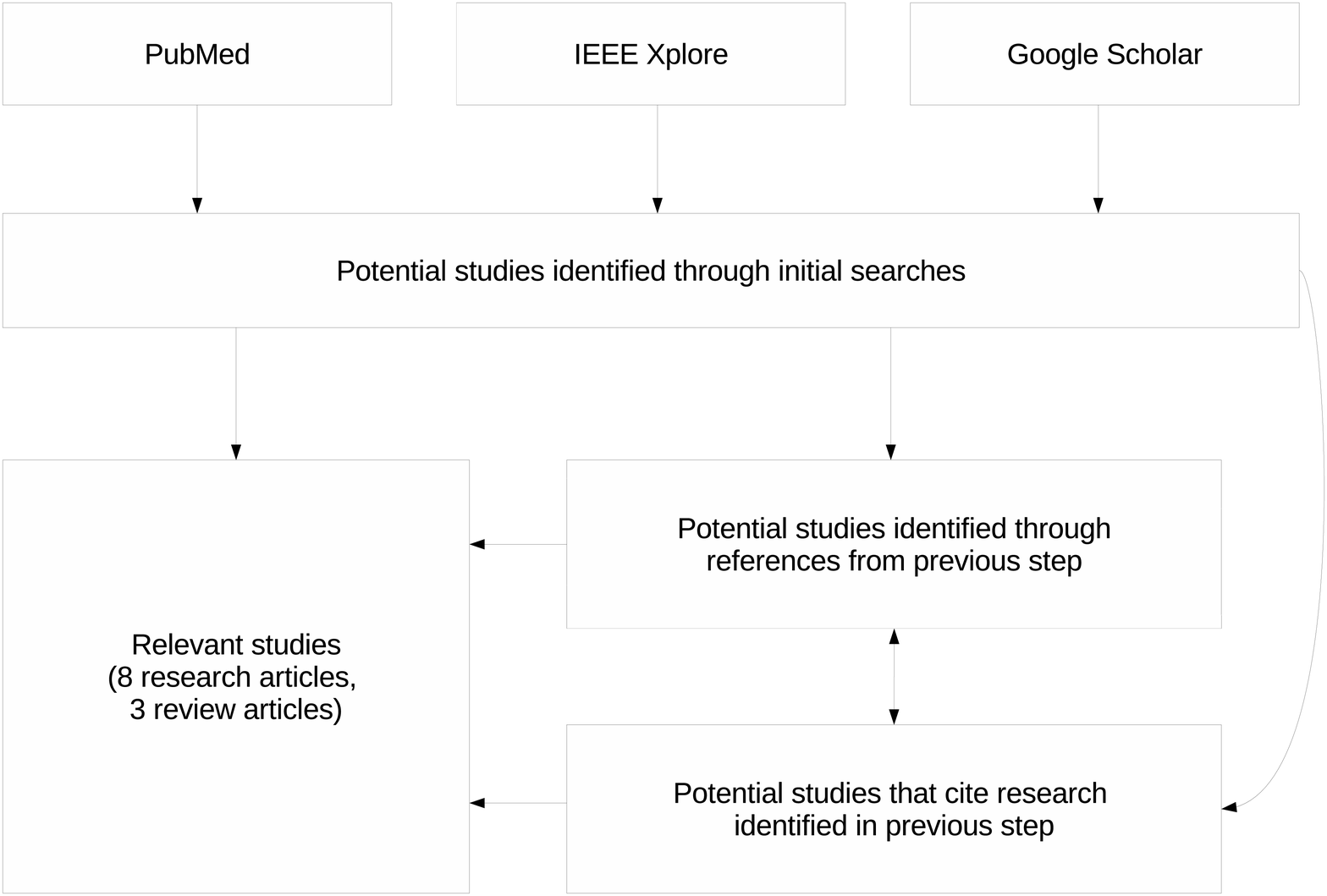}
  \caption{An overview of the study selection process.}
  \label{fig:methods_process}
\end{figure}

Exceptions to these criteria were made for existing surveys, literature reviews and meta-analyses that were directly concerned in a similar prediction problem. 
An overview of the study selection process is presented in figure \ref{fig:methods_process}. 
Studies fulfilling the criteria 
were searched initially from two databases, PubMed\footnote{https://www.ncbi.nlm.nih.gov} 
and IEEE Xplore\footnote{http://ieeexplore.ieee.org}, in addition to extensive searches on Google Scholar\footnote{https://scholar.google.com}. 
In the initial search phase, potential studies were mapped based on the titles of studies. 
PubMed search term was ((bmi[Title] OR (body mass index[Title])) AND (obesity[Title] OR obese[Title] OR overweight[Title]) 
AND (prediction[Title/Abstract] OR predicting[Title/Abstract])). This search yielded 63 results in total, of which four were identified 
as potential for inclusion. For IEEE Xplore, ''obesity'' search term was used with conference articles, journals and magazines starting from 1994 
included in the search. The search yielded 634 results with two potential studies. An additional set of potential studies were then searched using Google Scholar,  
by using variations of the search terms described before. These initial searches were made in July of 2017. 
After this initial seed set of potential studies was identified, references in studies and newer research that cite the studies were 
mapped for finding additional potentially relevant studies. Some of the relevant studies were identified already during these steps. 

A second search phase is indicated by a two-way arrow in figure \ref{fig:methods_process}. This searching through references of the potential 
studies and research that cited the potential studies continued until no new potential studies emerged. 
Each search cycle produced new material to the pool of potentially relevant studies.  
The idea and assumption behind this approach was that the relevant studies are most likely referenced in recent (2010--2018) research articles, 
so the search focused on the references of recent studies that were identified as potential for inclusion. 
This search phase lasted until August of 2018. 

\section{Results} \label{sec:results}

To our knowledge, this is the first study collecting the existing research that focuses on the predictive modeling approach in prediction of 
overweight or obesity. Three surveys, literature reviews and meta-analyses from years 2010--2015 closely related to the problem of overweight/obesity prediction are 
presented and discussed in section \ref{subsec:literature_reviews_and_meta_analyses}. 

Six out of eight studies from years 2004 to 2016 presented in section \ref{subsec:predictive_modeling_approaches} 
employ logistic regression in their models. Only two studies 
\cite{zhang:_comparing_data_mining_methods_with_logistic_regression_in_childhood_obesity_prediction,dugan:_machine_learning_techniques_for_prediction_of_early_childhood_obesity}
explore more complex models such as decision trees, Bayesian and neural networks, suppport vector machines, that have greater capacity to learn associations from data. 
Three of the studies predict 
only obesity and four only overweight including obesity. One study \cite{cheung:_a_longitudinal_study_of_pediatric_bmi_values_predicted_health_in_middle_age}
predicts separately two cases: overweight (including obesity) and obesity. 
The age range to predict overweight/obesity status in the studies was from two years to 33 years, 
with six studies predicting status in children under ten years of age. In addition to weight and height information of the child, 
number of features used in prediction varied from zero to five. The only exception \cite{dugan:_machine_learning_techniques_for_prediction_of_early_childhood_obesity} 
employed over 160 predictors in their prediction models.

\subsection{Surveys, literature reviews and meta-analyses} \label{subsec:literature_reviews_and_meta_analyses}

A survey of data mining methods used in the field was conducted by Adnan et al. (2010) \cite{hariz:_a_survey_on_utilization_of_data_mining_for_childhood_obesity_prediction}. 
The study summarized some of the research in the area and focused on describing three methods used in three different studies: 
neural networks, na\"{i}ve Bayes classifiers and decision trees. Each of the methods were described as having their own strengths and weaknesses. 
The conclusion was that to improve prediction results, further improvements of the techniques are necessary. The authors 
planned to continue their work by combining different existing methods to form a single better performing hybrid method.

Infancy weight gain was used as a predictor for childhood obesity in a meta-analysis presented by Druet et al. (2012)
\cite{druet:_prediction_of_childhood_obesity_by_infancy_weight_gain}. 
The meta-analysis reported a consistent positive association of infancy weight gain to subsequent obesity.
The study included ten cohort studies from which full data of  
three cohorts ($N=8236$) were used in forming models for overweight and obesity prediction. 
A second sample of the same size was used for validating the results. 
The prediction model employed stepwise multivariable logistic regression, and used mother's BMI, 
child's birthweight and sex in addition to weight gain information from birth to age of one year for childhood obesity prediction. 
The model was reported to show moderate predictive ability 
with area under curve (AUC) value of 0.77. When using a risk score threshold that puts 10 \% of the population above the threshold, 
the model had a sensitivity of 58.6 \% and specificity of 90.9 \%. A similar model was also created for childhood overweight prediction 
and reported AUC value was 0.76.

An extensive systematic review and meta-analysis of existing studies was presented by Simmonds et al. (2015)
\cite{simmonds:_the_use_of_measures_of_obesity_in_childhood_for_predicting_obesity} to examine the use of different 
measures of obesity in childhood for predicting obesity and development of obesity-related diseases. 
The study found that the predictive accuracy of childhood obesity for predicting adult obesity had a sensitivity of 30 \% 
and a specificity of 98 \% and was described as moderate. The study concluded that childhood BMI is not an effective 
predictor for obesity or disease in adulthood, since most obese adults were not obese in childhood. 
However, no evidence was found in the study to support any other single measure over BMI.

\subsection{Predictive modeling approaches} \label{subsec:predictive_modeling_approaches}

An overview of the studies employing predictive modeling is presented in table \ref{t:overview}. The studies might present also models that have not been validated with 
an independent data set, i.e. be partly explanatory in their approach. 
\textit{Predicted status} (overweight/obesity) as well as other information is listed only if the results were validated independently. 
\textit{Number of additional features used} exclude sex, weight, height and BMI of the child as well as overweight/obesity labels. 
If the study had separate models for females and males, best results (AUC, sensitivity and specificity) are presented for both. 
Two of the studies 
\cite{zhang:_comparing_data_mining_methods_with_logistic_regression_in_childhood_obesity_prediction,graversen:_prediction_of_adolescent_and_adult_adiposity_outcomes} 
presented different models for predicting 
overweight based on data collected up to three points of time. The best result for each of these models is also presented in the table. 
In addition to these, \cite{cheung:_a_longitudinal_study_of_pediatric_bmi_values_predicted_health_in_middle_age} presented separate models for 
obesity and overweight prediction. Again, the best results for both models are presented. 

\begin{table}[htp]
{\setstretch{1.0}
\resizebox{\columnwidth}{!}{%
\begin{threeparttable}[]
\begin{tabular}{lllllllll}
\textbf{Study}                                                                                                                                                              & \begin{tabular}[c]{@{}l@{}}Cheung et al.\\(2004)\cite{cheung:_a_longitudinal_study_of_pediatric_bmi_values_predicted_health_in_middle_age}\end{tabular}                                                                                                                                                             & \begin{tabular}[c]{@{}l@{}}Zhang et al.\\(2009)\cite{zhang:_comparing_data_mining_methods_with_logistic_regression_in_childhood_obesity_prediction}\end{tabular}                                                                      & \begin{tabular}[c]{@{}l@{}}Dugan et al.\\(2015)\cite{dugan:_machine_learning_techniques_for_prediction_of_early_childhood_obesity}\end{tabular}                                                                                                                               & \begin{tabular}[c]{@{}l@{}}Morandi et al.\\(2012)\cite{morandi:_estimation_of_newborn_risk_for_child_or_adolescent_obesity}\end{tabular}                                                                                                                               & \begin{tabular}[c]{@{}l@{}}Santorelli et al.\\(2013)\cite{santorelli:_developing_prediction_equations_and_a_mobile_phone_application}\end{tabular}                                                  & \begin{tabular}[c]{@{}l@{}}Weng et al.\\(2013)\cite{weng:_estimating_overweight_risk_in_childhood_from_predictors_during_infancy}\end{tabular}                                                                                                                                                       & \begin{tabular}[c]{@{}l@{}}Redsell et al.\\(2016)\cite{redsell:_validation_optimal_threshold_determination_and_clinical_utility}\end{tabular}                                                                                                                                                     & \begin{tabular}[c]{@{}l@{}}Graversen et al.\\(2015)\cite{graversen:_prediction_of_adolescent_and_adult_adiposity_outcomes}\end{tabular}                                                                       \\ \hline
\textbf{Method(s)}                                                                                                                                                            & \begin{tabular}[c]{@{}l@{}}ROC threshold\\ cut-off\end{tabular}                                                          & \begin{tabular}[c]{@{}l@{}}logistic regression,\\ decision tree,\\ association rules,\\ neural network,\\ linear SVM,\\ RBF SVM,\\ Bayesian network,\\ naïve Bayes\end{tabular} & \begin{tabular}[c]{@{}l@{}}random tree,\\ random forest,\\ ID3 decision tree,\\ J48 decision tree,\\ naïve Bayes,\\ Bayesian network\end{tabular} & \begin{tabular}[c]{@{}l@{}}stepwise\\ logistic regression\end{tabular}                                                                              & \begin{tabular}[c]{@{}l@{}}stepwise\\ logistic regression\end{tabular}                                      & \begin{tabular}[c]{@{}l@{}}stepwise\\ logistic regression\end{tabular}                                                                                                   & \begin{tabular}[c]{@{}l@{}}stepwise\\ logistic regression\end{tabular}                                                                                                    & logistic regression                                                                              \\ \hline
\textbf{\begin{tabular}[c]{@{}l@{}}Predicted\\ status\end{tabular}}                                                                                                         & \begin{tabular}[c]{@{}l@{}}overweight\\ (incl. obesity)\\ and obesity\end{tabular}                            & \begin{tabular}[c]{@{}l@{}}overweight\\ (incl. obesity)\end{tabular}                                                                                                              & obesity                                                                                                                                             & obesity                                                                                                                                               & obesity                                                                                                       & \begin{tabular}[c]{@{}l@{}}overweight\\ (incl. obesity)\end{tabular}                                                                                                       & \begin{tabular}[c]{@{}l@{}}overweight\\ (incl. obesity)\end{tabular}                                                                                                        & \begin{tabular}[c]{@{}l@{}}overweight\\ (incl. obesity)\end{tabular}    \\ \hline
\textbf{\begin{tabular}[c]{@{}l@{}}Age(s) to\\ predict\end{tabular}}                                                                                                     & 33 years                                                                                                                     & 3 years                                                                                                                                                                             & 2-10 years\tnote{1}                                                                                                                                           & 7 years                                                                                                                                                 & 2 years                                                                                                         & 3 years                                                                                                                                                                      & 5 years                                                                                                                                                                       & 13-16 years\tnote{2}                                                                                          \\ \hline
\textbf{\begin{tabular}[c]{@{}l@{}}Predicts\\ status from\\ child's\\ data\\ recorded\\ up to\end{tabular}}                                                                              & \begin{tabular}[c]{@{}l@{}}11 years\end{tabular}                                                  & \begin{tabular}[c]{@{}l@{}}6 weeks,\\ 8 months,\\ 2 years\end{tabular}                                                                                                          & \textless 2 years                                                                                                                                 & birth                                                                                                                                               & \begin{tabular}[c]{@{}l@{}}9 months,\\ 12 months\end{tabular}                                   & 1 year                                                                                                                                                                   & 1 year                                                                                                                                                                    & \begin{tabular}[c]{@{}l@{}}birth,\\ 5 years,\\ 8 years\end{tabular}                              \\ \hline
\textbf{\begin{tabular}[c]{@{}l@{}}Weight and\\ height\\ information\\ used\end{tabular}}                                                                                   & BMI                                                                                                                      & \begin{tabular}[c]{@{}l@{}}birth weight,\\ weight change,\\ height,\\ BMI\end{tabular}                                                                                          & \begin{tabular}[c]{@{}l@{}}height and\\ weight\end{tabular}                                                                                       & birth weight                                                                                                                                        & \begin{tabular}[c]{@{}l@{}}birth weight and\\ weight change\end{tabular}                                    & \begin{tabular}[c]{@{}l@{}}birth weight and\\ weight change\end{tabular}                                                                                                 & \begin{tabular}[c]{@{}l@{}}birth weight and\\ weight change\end{tabular}                                                                                                  & \begin{tabular}[c]{@{}l@{}}birth weight\\ and BMI\end{tabular}                                   \\ \hline
\textbf{\begin{tabular}[c]{@{}l@{}}Number of\\ additional\\ features used\end{tabular}} 									    & 0                                                                                                                        & \begin{tabular}[c]{@{}l@{}}0 (6 weeks),\\ 0 (8 months),\\ 1 (2 years)\end{tabular}                                                                                              & Over 160                                                                                                                                        & 5                                                                                                                                                   & 0                                                                                                           & 4                                                                                                                                                                        & 4                                                                                                                                                                         & 1                                                                                                \\ \hline
\textbf{\begin{tabular}[c]{@{}l@{}}Additional\\ features\end{tabular}}                                                                                                      & n/a                                                                                                                      & time of gestation                                                                                                                                                               & multiple\tnote{3}                                                                                                                                          & \begin{tabular}[c]{@{}l@{}}maternal BMI,\\ paternal BMI,\\ number of household\\ members,\\ maternal occupation,\\ gestational smoking\end{tabular} & n/a                                                                                                         & \begin{tabular}[c]{@{}l@{}}maternal pre-pregnancy\\ weight status,\\ paternal BMI,\\ maternal smoking\\ during pregnancy,\\ breastfeeding in the\\ first year\end{tabular} & \begin{tabular}[c]{@{}l@{}}maternal pre-pregnancy\\ weight status,\\ paternal BMI,\\ maternal smoking\\ during pregnancy,\\ breastfeeding in \\ the first year\end{tabular} & maternal BMI                                                                                     \\ \hline
\textbf{\begin{tabular}[c]{@{}l@{}}Validation\\ AUC (best)\end{tabular}}                                                                                                    & n/a                                                                                                                      & n/a                                                                                                                                                                             & n/a                                                                                                                                               & 0.73                                                                                                                                                & \begin{tabular}[c]{@{}l@{}}0.850 (9 months)\\  0.886 (12 months)\end{tabular}                                                                                                       & 0.755                                                                                                                                                                    & 0.79                                                                                                                                                                      & n/a                                                                                              \\ \hline
\textbf{\begin{tabular}[c]{@{}l@{}}Validation\\ sensitivity\\ (best)\end{tabular}}                                                                                          & \begin{tabular}[c]{@{}l@{}}75.7 \% (obesity)\tnote{*}\\ 71.7 \% (obesity)\tnote{**}\\ 69.8 \% (overweight)\tnote{*}\\ 65.6 \% (overweight)\tnote{**}\end{tabular}                                        & \begin{tabular}[c]{@{}l@{}}11.2 \% (6 weeks)\\ 35.5 \% (8 months)\\ 54.7 \% (2 years)\end{tabular}                                                                                                             & 89 \%                                                                                                                                             & n/a                                                                                                                                                 & n/a                                                                                                         & 76.9 \%                                                                                                                                                                  & 53 \%                                                                                                                                                                     & \begin{tabular}[c]{@{}l@{}}24.0 \% (birth)\tnote{*}\\ 17.4 \% (birth)\tnote{**}\\ 38.9 \% (5 years)\tnote{*}\\ 28.2 \% (5 years)\tnote{**}\\ 49.2 \% (8 years)\tnote{*}\\38.7 \%(8 years)\tnote{**} \end{tabular}            \\ \hline
\textbf{\begin{tabular}[c]{@{}l@{}}Validation\\ specificity\\ (best)\end{tabular}}                                                                                          & \begin{tabular}[c]{@{}l@{}}69.7 \% (obesity)\tnote{*}\\ 72.4 \% (obesity)\tnote{**}\\ 63.6 \% (overweight)\tnote{*}\\ 68.6 \% (overweight)\tnote{**}\end{tabular}                                        & \begin{tabular}[c]{@{}l@{}}96.0 \% (6 weeks)\\ 91.5 \% (8 months)\\ 93.1 \% (2 years)\end{tabular}                                                                                                             & 83 \%                                                                                                                                             & n/a                                                                                                                                                 & n/a                                                                                                         & 66.5 \%                                                                                                                                                                  & 71 \%                                                                                                                                                                     & \begin{tabular}[c]{@{}l@{}}92.1 \% (birth)\tnote{*}\\ 91.7 \% (birth)\tnote{**}\\ 94.4 \% (5 years)\tnote{*}\\ 94.2 \% (5 years)\tnote{**}\\ 96.0 \% (8 years)\tnote{*}\\96.7 \%(8 years)\tnote{**}\end{tabular}             \\ \hline
\textbf{\begin{tabular}[c]{@{}l@{}}N\\ (training data)\end{tabular}}                                                                                                        & 4136                                                                                                                     & $\sim$11070\tnote{4}                                                                                                                                                                    & \begin{tabular}[c]{@{}l@{}}$\sim$6767 per fold\\ (10-fold cv)\end{tabular}                                                        & 4032                                                                                                                                                & \begin{tabular}[c]{@{}l@{}}1528 (9 months)\\ 731 (12 months)\end{tabular}                                                                                                 & 8299                                                                                                                                                                     & \begin{tabular}[c]{@{}l@{}}8299\end{tabular}                                                                                                      & 4111                                                                                             \\ \hline
\textbf{\begin{tabular}[c]{@{}l@{}}N\\ (validation/\\ testing data)\end{tabular}}                                                                                           & 4231                                                                                                                     & $\sim$5400-5600\tnote{5}                                                                                                                                                               & \begin{tabular}[c]{@{}l@{}}$\sim$752 per fold\\ (10-fold cv)\end{tabular}                                                                         & 1503 and 1032                                                                                                                                       & \begin{tabular}[c]{@{}l@{}}880 (9 months)\\ 867 (12 months)\end{tabular}                                                                                              & 1715                                                                                                                                                                     & 980                                                                                                                                                                       & 5414                                                     \\ \hline                                     
\end{tabular}
\begin{tablenotes}
     \item[*] Females.
     \item[**] Males.
     \item[1] If a child was obese at any point of time between ages two to ten years, he/she was labeled as obese.
     \item[2] If a child was overweight at any point of time between ages 13 to 16 years, he/she was labeled as overweight.
     \item[3] 167 variables in total, but they are not listed in detail. A questionnaire form used to collect the information is presented.
     \item[4] 67 \% of total data.
     \item[5] 33 \% of total data from 16523 equals 5453 observations. However, the confusion matrices in figures 3 and 4 in the study include 5618 observations.
   \end{tablenotes}
\end{threeparttable}
}
\caption{Overview of the studies using predictive modeling methods for overweight/obesity prediction based on childhood data.}\label{t:overview}
}
\end{table}

Data collected at childhood was used by Cheung et al. (2004) \cite{cheung:_a_longitudinal_study_of_pediatric_bmi_values_predicted_health_in_middle_age} 
to predict obesity and overweight at age 33 years and self-reported disease history at age 42 years. 
ROC analysis was used to define optimal BMI risk thresholds at ages seven, 11 and 16. The cohort data was split in half, with the first half used for the ROC analysis and 
the second one for validating the cut-off BMI thresholds. For the validation data ($N=4231$) at age 11 years, sensitivity and specificity values for obesity status prediction were 
respectively 71.7 \% and 72.4 \% for males. For overweight prediction, the same values were 65.6 \% and 68.6 \%. Obesity prediction for females achieved sensitivity of 75.7 \% 
and specificity of 69.7 \%, while the same values for overweight prediction were 69.8 \% and 63.6 \%.

Six independent data mining methods were investigated and compared to the conventional logistic regression approach by Zhang et al. (2009) 
\cite{zhang:_comparing_data_mining_methods_with_logistic_regression_in_childhood_obesity_prediction}. The study included 
C4.5 decision tree, Bayesian networks, na\"{i}ve Bayes, association rules, neural networks and support vector machines (SVM). 
The task was to predict overweight status at age three years based on child's data ($N=16523$) collected up to six weeks, eight months and two years respectively. 
67 \% of the data was used for training the models and 33 \% for testing. 
Children's height and weight data were used in the models. Time of gestation was used as an additional feature when predicting overweight status from the age of two years.
The results showed that the prediction accuracy improved with the methods used when compared to the logistic regression approach. 
The best performing algorithms were SVM with radial basis function (RBF) kernel and Bayesian methods, specifically na\"{i}ve Bayes. 
At age two years, the sensitivity and specificity of NB were 54.7 \% and 93.1 \% respectively, while the same values for SVM 
were 60.0 \% and 79.6 \%.
The conclusion was that to improve overweight prediction rates more features may need to be recorded and used for prediction.

The aim of Dugan et al. (2015)\cite{dugan:_machine_learning_techniques_for_prediction_of_early_childhood_obesity} was to improve the work presented in 
\cite{zhang:_comparing_data_mining_methods_with_logistic_regression_in_childhood_obesity_prediction} by considering a significantly 
extended set of predictors. A partly different set of machine learning algorithms were also explored in the study. 
They included random tree, random forest, 
ID3 and C4.5 decision trees in addition to na\"{i}ve Bayes and Bayesian networks. The study employed in total 167 predictors collected 
through questionnaires filled by parents and physicians. The data ($N=7519$) were collected before the children's second birthday. 
10-fold cross-validation was used for validating the results. 
If the child was obese at any point after her/his second birthday to age of 10 years, the child was labeled as obese. The prediction task 
consisted of predicting the child's future obesity status after the second birthday. 
Two of the algorithms, ID3 and NB, performed slightly better when predictors described as ''noisy'' were removed from the data. 
The ID3 model used 87 and the NB model 107 out of 167 predictors. Other methods employed in the study performed best when all the available 
predictors were utilized. 
The study singled out two algorithms as best working with this data set: ID3 decision tree with sensitivity and 
specificity of 89 \% and 83 \% respectively, along with random tree's performance metrics of 88 \% and 80 \%.

Stepwise logistic regression analysis was used to form predictive models by Morandi et al. (2012) \cite{morandi:_estimation_of_newborn_risk_for_child_or_adolescent_obesity} to 
predict obesity status at age of seven years with data available at birth. The final model included maternal and paternal BMI, number of household members, maternal occupation, 
gestational smoking as well as birth weight. 
First cohort data ($N=4032$) was used in training the model, while validation of the results was performed using two independent cohorts ($N_1=1503$, $N_2=1032$). 
Reported AUC values for the two validation sets were 0.70 and 0.73 respectively. For the first validation cohort, gestational smoking and number of household members information 
had to be omitted from the model. The study also experimented with adding a genetic score information to the models, but that did not provide any significant improvement in terms of 
predictive power. 

Three logistic regression models were formed with stepwise method by Santorelli et al. (2013)\cite{santorelli:_developing_prediction_equations_and_a_mobile_phone_application}, where 
the models were applied in a mobile application aimed at parents. The models predicted the risk of childhood obesity at age of two years. The plan presented in the study was 
that the parents use the application when their infant's age reaches six, nine and twelve months, with separate models presented for each case. Input features included sex, birth weight and weight gain. 
Two datasets were used. The first one was used to form the three models ($N_1=1022$, $N_2=1528$, $N_3=731$) and validated using an internal bootstrap validation. 
The second dataset was used as a separate external testing data. The second and third model were validated using this data ($N_2=880$, $N_3=867$), with the reported AUC values 
for the two models being 0.85 and 0.89 respectively. More detailed sensitivity and specificity values were also reported for different model configurations, but only for the 
training data. The mobile application has since been discontinued, and no published research exists on the usage or effectiveness of the application 
\cite{butler:_childhood_obesity__how_long_should_we_wait_to_predict_weight}.

Weng et al. (2013) \cite{weng:_estimating_overweight_risk_in_childhood_from_predictors_during_infancy} formed a childhood overweight prediction model (IROC)
using stepwise logistic regression for predicting obesity status at age of three years. The cohort data was randomly divided into training ($N=8299$) and testing ($N=1715$) sets. 
The data included 33 potential predictor variables. From these, seven significant input features were identified: sex, birth weight, weight gain in first year, 
maternal pre-pregnancy weight status, paternal BMI, maternal smoking during pregnancy and breastfeeding in the first year. 
The study reported a moderately good predictive ability, with sensitivity value of 0.769, specificity of 0.665 and AUC value 0.755 for the 
testing data.

The IROC algorithm \cite{weng:_estimating_overweight_risk_in_childhood_from_predictors_during_infancy} was further validated by Redsell et al. (2016) 
\cite{redsell:_validation_optimal_threshold_determination_and_clinical_utility} with an additional independent dataset ($N=980$), predicting overweight 
status at age of five years. Four models were formed for prediction. The first one (clinical model) used the original algorithm directly and assigned null values 
to missing data. The reported AUC values were 0.67 when using the International obesity taskforce overweight criteria 
\cite{cole:_establishing_a_standard_definition_of_childhood_obesity_worldwide} and 0.65 when using the UK 1990 overweight criteria \cite{cole:_body_mass_index_reference_curves_for_the_uk}. 
The second one (recalibrated model) used multivariate logistic regression to generate and recalibrate the model to reflect the demographics of 
the new validation data (AUC values of 0.70 and 0.67 were reported). The third one (imputed model) used multiple imputation to generate ten copies of the existing data set
to predict missing risk factor from multivariate models (AUC values 0.79 and 0.73). 
The fourth one (recalibrated imputed model) applied the recalibrated algorithm to the imputed data (AUC values 0.93 and 0.90). 

Graversen et al. (2015) \cite{graversen:_prediction_of_adolescent_and_adult_adiposity_outcomes} developed logistic regression models to predict adolescent overweight, 
adult overweight and adult obesity. Input features for the models included maternal BMI, birth weight and early childhood BMI. First dataset ($N=4111$) was used 
to form the models. Performance of the models was validated with an internal bootstrap validation as well as with an external independent dataset ($N=5414$), where 
the prevalence of overweight was much higher. Only the model for adolescent overweight prediction was validated with the external data. 
The study reported results for adolescent overweight prediction from data collected at birth, up to age of five years and 
up to age of eight years. Also, different thresholds of percentage of children labeled as ''at risk'' of overweight were explored. 
With the external validation dataset, sensitivity and specificity for adolescent overweight prediction at age of five years for females 
were 38.9 \% and 94.4 \% respectively when the threshold of being at risk was set to upper 10 \%. For males these measures were 28.2 \% and 94.2 \%.

\section{Conclusions} \label{sec:conclusions}

This review explored the existing research on overweight and obesity prediction. While various explanatory models have 
been studied and employed extensively in the research area, utilization of predictive modeling methods of machine 
learning remain partly unexplored in the field.

The studies using explanatory modeling do not validate the formed models with a separate test data. 
Instead they examine how well the whole data fits to the model. 
Generalization refers to how well the model trained on the training data set predicts the output for new instances, 
and it is an integral part of the machine learning and predictive modeling approach 
\cite{alpaydin:_introduction_to_machine_learning, bishop:_pattern_recognition_and_machine_learning}. 
We argue that if the model is aimed for prediction, the model should always be validated with independent data to get more reliable performance estimates.

In terms of predictive power, best performing models in the study either made the prediction quite late 
or had a relatively short period between the prediction and the outcome. \cite{graversen:_prediction_of_adolescent_and_adult_adiposity_outcomes} 
had a very high specificity (96 \%) in predicting overweight at adolescence in girls, 
using data recorded up to eight years. Other moderately successful models, 
\cite{santorelli:_developing_prediction_equations_and_a_mobile_phone_application,dugan:_machine_learning_techniques_for_prediction_of_early_childhood_obesity}, 
and ''two years to three years'' model by 
\cite{zhang:_comparing_data_mining_methods_with_logistic_regression_in_childhood_obesity_prediction} 
only had a short time period to cover. So far there has been no evidence of success in employing any of the presented or other models in clinical use 
\cite{butler:_childhood_obesity__how_long_should_we_wait_to_predict_weight}.

The constantly growing size of data sets will enable the use of even more powerful machine learning methods and, 
moreover, sophisticated analysis of the obtained models. In the present context, one completely unexplored 
approach is, for example, the recurrent neural networks \cite{hochreiter:_lstm}. They could provide a powerful method for predicting 
overweight and obesity development in the later life, because they are inherently designed for time series prediction 
tasks (e.g. \cite{choi:_using_recurrent_neural_network_models_for_early_detection_of_heart_failure_onset}). 
It is, however, important to collect a sufficient amount of data before training complex models, 
such as, neural networks.

Besides being powerful methodology for building prediction models, more effort could also be directed to machine 
learning based hypothesis generation by employing large data sets, high-performance computing and machine learning 
algorithms. This can lead to finding of unsuspected information and predictors from the growing data.

A critical issue to be considered when applying machine learning in overweight and obesity prediction tasks is, 
however, the risk of chance findings. In order to minimize the risk  of chance models or predictors, it is highly 
important to develop and apply strategies, such as data randomization \cite{ojala:_permutation_tests_for_studying_classifier_performance}, 
for confirming significance of the obtained models and relevance of the identified predictors.

\section*{Acknowledgements}\label{sec:acknowledgements}

We thank Tekes / Business Finland for financial support and Richard Allmendinger for his comments on the initial draft article. 
The funding source did not have any other involvement in the study.

\section*{Summary points}

What was already known on the topic
\begin{itemize}
 \item A lot of research has been done on predicting upcoming obesity or overweight.
 \item None of the prediction models presented in previous research have been successfully utilized in practice. 
\end{itemize}

What this study added to our knowledge
\begin{itemize}
 \item Although prediction in domain of overweight/obesity has been studied extensively, there are not many studies that separate the training 
 and validation of the model in a way expected in predictive modeling. 
 \item Highest performing prediction models are either only predicting near future overweight/obesity status or make their prediction relatively late. 
 \item More complex models with greater learning capacity have been employed in only a limited set of studies. Future research possibilities in the area 
exist in examining the potential of recurrent models and considering further model building practices such as predictor selection and significance of 
prediction.
\end{itemize}

\bibliographystyle{elsarticle-num}
\bibliography{literature_review_srcs}

\begin{thebibliography}{10}
\expandafter\ifx\csname url\endcsname\relax
  \def\url#1{\texttt{#1}}\fi
\expandafter\ifx\csname urlprefix\endcsname\relax\def\urlprefix{URL }\fi
\expandafter\ifx\csname href\endcsname\relax
  \def\href#1#2{#2} \def\path#1{#1}\fi

\bibitem{the_gbd_2015_obesity_collaborators:_health_effects_of_overweight_and_obesity}
{The GBD 2015 Obesity Collaborators}, Health effects of overweight and obesity
  in 195 countries over 25 years, New England Journal of Medicine 377~(1)
  (2017) 13--27.
\newblock \href {http://dx.doi.org/10.1056/NEJMoa1614362}
  {\path{doi:10.1056/NEJMoa1614362}}.

\bibitem{onis:_development_of_a_who_growth_reference}
M.~d. Onis, A.~W. Onyango, E.~Borghi, A.~Siyam, C.~Nishida, J.~Siekmann,
  Development of a who growth reference for school-aged children and
  adolescents, Bulletin of the World Health Organization 85 (2007) 660--667.
\newblock \href {http://dx.doi.org/10.1590/S0042-96862007000900010}
  {\path{doi:10.1590/S0042-96862007000900010}}.

\bibitem{saari:_new_finnish_growth_references_for_children_and_adolescents}
A.~Saari, U.~Sankilampi, M.-L. Hannila, V.~Kiviniemi, K.~Kesseli, L.~Dunkel,
  New finnish growth references for children and adolescents aged 0 to 20
  years: Length/height-for-age, weight-for-length/height, and body mass
  index-for-age, Annals of Medicine 43~(3) (2011) 235--248.
\newblock \href {http://dx.doi.org/10.3109/07853890.2010.515603}
  {\path{doi:10.3109/07853890.2010.515603}}.

\bibitem{cole:_extended_international_iotg_bmi_cutoffs}
T.~J. Cole, T.~Lobstein, Extended international (iotf) body mass index cut-offs
  for thinness, overweight and obesity, Pediatric Obesity 7~(4) (2012)
  284--294.
\newblock \href {http://dx.doi.org/10.1111/j.2047-6310.2012.00064.x}
  {\path{doi:10.1111/j.2047-6310.2012.00064.x}}.

\bibitem{lynch:_development_of_distinct_bmi_trajectories_among_children_before_age_5_years}
B.~A. Lynch, L.~J.~F. Rutten, J.~O. Ebbert, S.~Kumar, B.~P. Yawn, D.~Jacobson,
  J.~S. Sauver, Development of distinct body mass index trajectories among
  children before age 5 years: A population-based study, Journal of Primary
  Care \& Community Health 8~(4) (2017) 278--284.
\newblock \href {http://dx.doi.org/10.1177/2150131917704326}
  {\path{doi:10.1177/2150131917704326}}.

\bibitem{cole:_establishing_a_standard_definition_of_childhood_obesity_worldwide}
T.~J. Cole, M.~C. Bellizzi, K.~M. Flegal, W.~H. Dietz, Establishing a standard
  definition of childhood obesity worldwide: International survey, BMJ
  320~(7244) (2000) 1240--1243.
\newblock \href {http://dx.doi.org/10.1136/bmj.320.7244.1240}
  {\path{doi:10.1136/bmj.320.7244.1240}}.

\bibitem{cole:_body_mass_index_reference_curves_for_the_uk}
T.~Cole, J.~Freeman, M.~Preece, Body mass index reference curves for the uk,
  1990., Archives of Disease in Childhood 73~(1) (1995) 25--29.
\newblock \href {http://dx.doi.org/10.1136/adc.73.1.25}
  {\path{doi:10.1136/adc.73.1.25}}.

\bibitem{zhang:_comparing_data_mining_methods_with_logistic_regression_in_childhood_obesity_prediction}
S.~Zhang, C.~Tjortjis, X.~Zeng, H.~Qiao, I.~Buchan, J.~Keane, Comparing data
  mining methods with logistic regression in childhood obesity prediction,
  Information Systems Frontiers 11~(4) (2009) 449--460.
\newblock \href {http://dx.doi.org/10.1007/s10796-009-9157-0}
  {\path{doi:10.1007/s10796-009-9157-0}}.

\bibitem{butler:_childhood_obesity__how_long_should_we_wait_to_predict_weight}
{\'E}.~M. Butler, J.~G. Derraik, R.~W. Taylor, W.~S. Cutfield, Childhood
  obesity: How long should we wait to predict weight?, Journal of pediatric
  endocrinology and metabolism 31~(5) (2018) 497--501.
\newblock \href {http://dx.doi.org/10.1515/jpem-2018-0110}
  {\path{doi:10.1515/jpem-2018-0110}}.

\bibitem{levine:_identifying_infants_at_risk_of_becoming_obese__can_we_and_should_we}
R.~Levine, D.~Dahly, M.~Rudolf, Identifying infants at risk of becoming obese:
  Can we and should we?, Public health 126~(2) (2012) 123--128.
\newblock \href {http://dx.doi.org/10.1016/j.puhe.2011.10.008}
  {\path{doi:10.1016/j.puhe.2011.10.008}}.

\bibitem{breiman:_statistical_modeling}
L.~Breiman, Statistical modeling: The two cultures, Statistical science 16~(3)
  (2001) 199--215.
\newblock \href {http://dx.doi.org/10.1214/ss/1009213726}
  {\path{doi:10.1214/ss/1009213726}}.

\bibitem{shmueli:_to_explain_or_to_predict}
G.~Shmueli, To explain or to predict?, Statistical science 25~(3) (2010)
  289--310.
\newblock \href {http://dx.doi.org/10.1214/10-STS330}
  {\path{doi:10.1214/10-STS330}}.

\bibitem{bishop:_pattern_recognition_and_machine_learning}
C.~M. Bishop, Pattern recognition and machine learning, Springer, 2006.

\bibitem{bzdok:_points_of_significance}
D.~Bzdok, N.~Altman, M.~Krzywinski, Points of significance: statistics versus
  machine learning, Nature Methods 15~(4) (2018) 233--234.
\newblock \href {http://dx.doi.org/10.1038/nmeth.4642}
  {\path{doi:10.1038/nmeth.4642}}.

\bibitem{sainani:_explanatory_versus_predictive_modeling}
K.~L. Sainani, Explanatory versus predictive modeling, PM\&R 6~(9) (2014)
  841--844.
\newblock \href {http://dx.doi.org/10.1016/j.pmrj.2014.08.941}
  {\path{doi:10.1016/j.pmrj.2014.08.941}}.

\bibitem{dugan:_machine_learning_techniques_for_prediction_of_early_childhood_obesity}
T.~M. Dugan, S.~Mukhopadhyay, A.~Carroll, S.~Downs, et~al., Machine learning
  techniques for prediction of early childhood obesity, Applied clinical
  informatics 6~(3) (2015) 506--520.
\newblock \href {http://dx.doi.org/10.4338/ACI-2015-03-RA-0036}
  {\path{doi:10.4338/ACI-2015-03-RA-0036}}.

\bibitem{cheung:_a_longitudinal_study_of_pediatric_bmi_values_predicted_health_in_middle_age}
Y.~B. Cheung, D.~Machin, J.~Karlberg, K.~S. Khoo, A longitudinal study of
  pediatric body mass index values predicted health in middle age, Journal of
  Clinical Epidemiology 57~(12) (2004) 1316 -- 1322.
\newblock \href {http://dx.doi.org/10.1016/j.jclinepi.2004.04.010}
  {\path{doi:10.1016/j.jclinepi.2004.04.010}}.

\bibitem{hariz:_a_survey_on_utilization_of_data_mining_for_childhood_obesity_prediction}
M.~H. B.~M. Adnan, W.~Husain, F.~Damanhoori, A survey on utilization of data
  mining for childhood obesity prediction, in: 8th Asia-Pacific Symposium on
  Information and Telecommunication Technologies, 2010, pp. 1--6.

\bibitem{druet:_prediction_of_childhood_obesity_by_infancy_weight_gain}
C.~Druet, N.~Stettler, S.~Sharp, R.~K. Simmons, C.~Cooper, G.~Davey~Smith,
  U.~Ekelund, C.~Lévy-Marchal, M.-R. J\"{a}rvelin, D.~Kuh, K.~K. Ong,
  Prediction of childhood obesity by infancy weight gain: an individual-level
  meta-analysis, Paediatric and Perinatal Epidemiology 26~(1) (2012) 19--26.
\newblock \href {http://dx.doi.org/10.1111/j.1365-3016.2011.01213.x}
  {\path{doi:10.1111/j.1365-3016.2011.01213.x}}.

\bibitem{simmonds:_the_use_of_measures_of_obesity_in_childhood_for_predicting_obesity}
M.~Simmonds, J.~Burch, A.~Llewellyn, C.~Griffiths, H.~Yang, C.~Owen, S.~Duffy,
  N.~Woolacott, The use of measures of obesity in childhood for predicting
  obesity and the development of obesity-related diseases in adulthood: a
  systematic review and meta-analysis., Health technology assessment
  (Winchester, England) 19~(43) (2015) 1 -- 336.
\newblock \href {http://dx.doi.org/10.3310/hta19430}
  {\path{doi:10.3310/hta19430}}.

\bibitem{graversen:_prediction_of_adolescent_and_adult_adiposity_outcomes}
L.~Graversen, T.~I. S{\o}rensen, T.~A. Gerds, L.~Petersen, U.~Sovio,
  M.~Kaakinen, A.~Sandbaek, J.~Laitinen, A.~Taanila, A.~Pouta, M.-R.
  J\"{a}rvelin, C.~Obel, Prediction of adolescent and adult adiposity outcomes
  from early life anthropometrics, Obesity 23~(1) (2015) 162--169.
\newblock \href {http://dx.doi.org/10.1002/oby.20921}
  {\path{doi:10.1002/oby.20921}}.

\bibitem{morandi:_estimation_of_newborn_risk_for_child_or_adolescent_obesity}
A.~Morandi, D.~Meyre, S.~Lobbens, K.~Kleinman, M.~Kaakinen, S.~L. Rifas-Shiman,
  V.~Vatin, S.~Gaget, A.~Pouta, A.-L. Hartikainen, J.~Laitinen, A.~Ruokonen,
  S.~Das, A.~A. Khan, P.~Elliott, C.~Maffeis, M.~W. Gillman, M.-R.
  J\"{a}rvelin, P.~Froguel, Estimation of newborn risk for child or adolescent
  obesity: Lessons from longitudinal birth cohorts, PLOS ONE 7~(11) (2012)
  1--9.
\newblock \href {http://dx.doi.org/10.1371/journal.pone.0049919}
  {\path{doi:10.1371/journal.pone.0049919}}.

\bibitem{santorelli:_developing_prediction_equations_and_a_mobile_phone_application}
G.~Santorelli, E.~S. Petherick, J.~Wright, B.~Wilson, H.~Samiei, N.~Cameron,
  W.~Johnson, Developing prediction equations and a mobile phone application to
  identify infants at risk of obesity, PLOS ONE 8~(8) (2013) 1--8.
\newblock \href {http://dx.doi.org/10.1371/journal.pone.0071183}
  {\path{doi:10.1371/journal.pone.0071183}}.

\bibitem{weng:_estimating_overweight_risk_in_childhood_from_predictors_during_infancy}
S.~F. Weng, S.~A. Redsell, D.~Nathan, J.~A. Swift, M.~Yang, C.~Glazebrook,
  Estimating overweight risk in childhood from predictors during infancy,
  Pediatrics 132~(2) (2013) e414--e421.
\newblock \href {http://dx.doi.org/10.1542/peds.2012-3858}
  {\path{doi:10.1542/peds.2012-3858}}.

\bibitem{redsell:_validation_optimal_threshold_determination_and_clinical_utility}
S.~A. Redsell, S.~Weng, J.~A. Swift, D.~Nathan, C.~Glazebrook, Validation,
  optimal threshold determination, and clinical utility of the infant risk of
  overweight checklist for early prevention of child overweight, Childhood
  Obesity 12~(3) (2016) 202--209.
\newblock \href {http://dx.doi.org/10.1089/chi.2015.0246}
  {\path{doi:10.1089/chi.2015.0246}}.

\bibitem{alpaydin:_introduction_to_machine_learning}
E.~Alpayd{\i}n, Introduction to machine learning: third edition, MIT Press,
  2014.

\bibitem{hochreiter:_lstm}
S.~Hochreiter, J.~Schmidhuber, Long short-term memory, Neural computation 9~(8)
  (1997) 1735--1780.
\newblock \href {http://dx.doi.org/10.1162/neco.1997.9.8.1735}
  {\path{doi:10.1162/neco.1997.9.8.1735}}.

\bibitem{choi:_using_recurrent_neural_network_models_for_early_detection_of_heart_failure_onset}
E.~Choi, A.~Schuetz, W.~F. Stewart, J.~Sun, Using recurrent neural network
  models for early detection of heart failure onset, Journal of the American
  Medical Informatics Association 24~(2) (2017) 361--370.
\newblock \href {http://dx.doi.org/10.1093/jamia/ocw112}
  {\path{doi:10.1093/jamia/ocw112}}.

\bibitem{ojala:_permutation_tests_for_studying_classifier_performance}
M.~Ojala, G.~C. Garriga, Permutation tests for studying classifier performance,
  Journal of Machine Learning Research 11~(Jun) (2010) 1833--1863.

\end{thebibliography}
\end{document}